\documentclass[11pt,a4paper]{article}
\usepackage[hyperref]{emnlp2020}
\usepackage{times}
\usepackage{latexsym}
\usepackage{paralist}

\usepackage{microtype}

\aclfinalcopy %

\usepackage[utf8]{inputenc}
\usepackage{color,soul}
\setul{0.45ex}{0.25ex}

\usepackage{makecell}
\usepackage{graphicx}
\usepackage{tabularx}

\usepackage{subcaption}

\usepackage{amsmath}
\usepackage{booktabs}
\usepackage{multirow}
\usepackage{graphicx}
\usepackage{array}
\usepackage{algorithm}
\usepackage{url}

\usepackage{listings}

\usepackage{placeins}

\usepackage{fancyhdr}
\usepackage{lipsum}
\usepackage{lipsum}
\usepackage{titlesec}

\titlespacing\section{0pt}{2pt plus 1pt minus 1pt}{2pt plus 0pt minus 1pt}
\titlespacing\subsection{0pt}{1pt plus 0pt minus 1pt}{0pt plus 0pt minus 0pt}
\titlespacing\subsubsection{0pt}{0pt plus 0pt minus 1pt}{0pt plus 0pt minus 0pt}
\titlespacing{\paragraph}{1pt}{0pt}{0pt}[0pt]

\usepackage{etoolbox}
\makeatletter
\preto{\@tabular}{\parskip=0pt}
\makeatother

\usepackage[font=small,skip=2pt]{caption}

\usepackage{enumitem}
\setlist[itemize]{leftmargin=*}

\title{\texttt{TextAttack}: A Framework for Adversarial Attacks, Data Augmentation, and Adversarial Training in NLP}

\author{John X. Morris\textsuperscript{1},
Eli Lifland\textsuperscript{1}, 
Jin Yong Yoo\textsuperscript{1}, 
Jake Grigsby\textsuperscript{1}, Di Jin\textsuperscript{2},  Yanjun Qi\textsuperscript{1} \\
\textsuperscript{1} Department of Computer Science, University of Virginia \\
\textsuperscript{2} Computer Science and Artificial Intelligence Laboratory, MIT\\
\{\href{mailto:jm8wx@virginia.edu}{jm8wx}, \href{mailto:yq2h@virginia.edu}{yq2h}\}@virginia.edu
\\
}

\date{}

\begin{document}
\maketitle

\begin{abstract}

While there has been substantial research using adversarial attacks to analyze NLP models, each attack is implemented in its own code repository. It remains challenging to develop NLP attacks and utilize them to improve model performance.
This paper introduces \texttt{TextAttack}, a Python framework for adversarial attacks, data augmentation, and adversarial training in NLP.
\texttt{TextAttack} builds attacks from four components: a goal function, a set of constraints, a transformation, and a search method.
\texttt{TextAttack}'s modular design enables researchers to easily construct attacks from combinations of novel and existing components. \texttt{TextAttack} provides implementations of 16 adversarial attacks from the literature and supports a variety of models and datasets, including BERT and other transformers, and all GLUE tasks.
\texttt{TextAttack} also includes data augmentation and adversarial training modules for using components of adversarial attacks to improve model accuracy and robustness. \texttt{TextAttack} is democratizing NLP: anyone can try data augmentation and adversarial training on any model or dataset, with just a few lines of code.
Code and tutorials are available at \href{https://github.com/QData/TextAttack}{https://github.com/QData/TextAttack}.

\end{abstract}
\section{Introduction}

Over the last few years, there has been growing interest in investigating the adversarial robustness of NLP models, including new methods for generating adversarial examples and better approaches to defending against these adversaries \cite{Alzantot2018GeneratingNL,Jin2019TextFooler,Kuleshov2018AdversarialEF,Li2019TextBuggerGA,Gao2018BlackBoxGO,iga-wang2019natural,Ebrahimi2017HotFlipWA,pso-zang-etal-2020-word,pruthi2019combating}. It is difficult to compare these attacks directly and fairly, since they are often evaluated on different data samples and victim models. Re-implementing previous work as a baseline is often time-consuming and error-prone due to a lack of source code, and precisely replicating results is complicated by small details left out of the publication. These barriers make benchmark comparisons hard to trust and severely hinder the development of this field.

\begin{figure}[]
\includegraphics[width=\columnwidth]{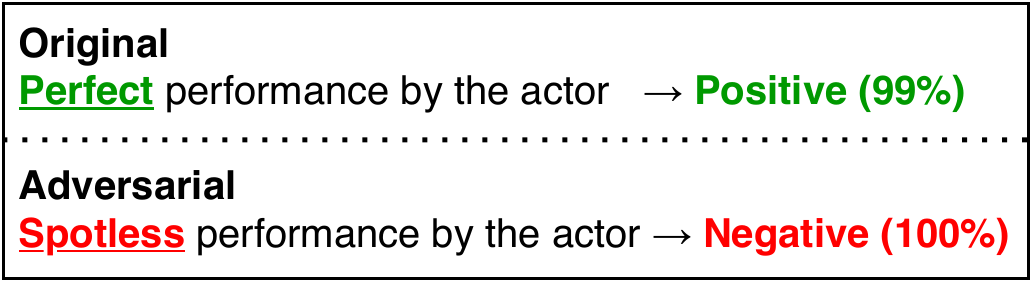}
\caption{Adversarial example generated using \citet{Jin2019TextFooler}'s \texttt{TextFooler} for a BERT-based sentiment classifier.  Swapping out "perfect" with synonym "spotless" completely changes the model's prediction, even though the underlying meaning of the text has not changed. \label{fig:ae-example}}
\end{figure}

\begin{figure*}[tb]
\centering
\includegraphics[width=\textwidth]{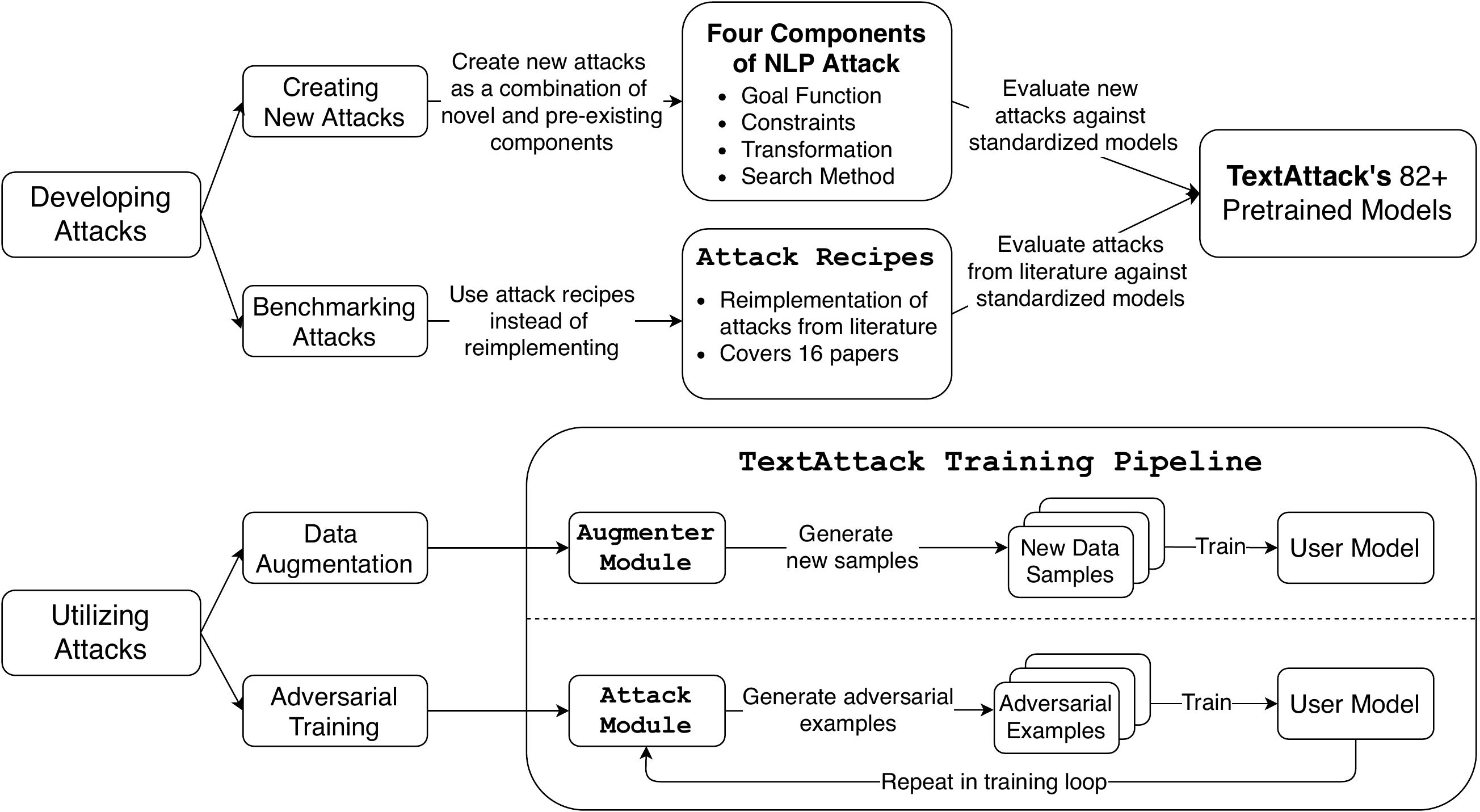}
\caption{Main features of \texttt{TextAttack}.\label{fig:overall-fig}}
\vspace{-3mm}
\end{figure*}

To encourage the development of the adversarial robustness field, we introduce \texttt{TextAttack}, a Python framework for adversarial attacks, data augmentation, and adversarial training in NLP.

To unify adversarial attack methods into one system, we decompose NLP attacks into four components: 
a goal function, a set of constraints, a transformation, and a search method. The attack attempts to perturb an input text such that the model output fulfills the goal function (i.e., indicating whether the attack is successful) and the perturbation adheres to the set of constraints (e.g., grammar constraint, semantic similarity constraint). A search method is used to find a sequence of transformations that produce a successful adversarial example.

This modular design enables us to easily assemble attacks from the literature while re-using components that are shared across attacks. \texttt{TextAttack} provides clean, readable implementations of 16 adversarial attacks from the literature. For the first time, these attacks can be benchmarked, compared, and analyzed in a standardized setting. \texttt{TextAttack}'s design also allows researchers to easily construct new attacks from combinations of novel and existing components. In just a few lines of code, the same search method, transformation and constraints used in \citet{Jin2019TextFooler}'s \texttt{TextFooler} can be modified to attack a translation model with the goal of changing every word in the output.

\texttt{TextAttack} is directly integrated with HuggingFace's \href{https://github.com/huggingface/transformers/}{transformers} and \href{https://github.com/huggingface/nlp/}{nlp} libraries. This allows users to test attacks on models and datasets. \texttt{TextAttack} provides dozens of pre-trained models (LSTM, CNN, and various transformer-based models) on a variety of popular datasets. Currently \texttt{TextAttack} supports a multitude of tasks including summarization, machine translation, and all nine tasks from the GLUE benchmark. \texttt{TextAttack} also allows users to provide their own models and datasets.

Ultimately, the goal of studying adversarial attacks is to improve model performance and robustness. To that end, \texttt{TextAttack} provides easy-to-use tools for data augmentation and adversarial training. \texttt{TextAttack}'s \texttt{Augmenter} class uses a transformation and a set of constraints to produce new samples for data augmentation. Attack recipes are re-used in a training loop that allows models to train on adversarial examples. These tools make it easier to train accurate and robust models.

Uses for \texttt{TextAttack} include\footnote{All can be done in $< 5$ lines of code. See \ref{appendix:five-lines-or-less}.}:
\begin{itemize}[topsep=0pt, partopsep=0pt]
  \setlength\itemsep{0em}
  \setlength{\parskip}{0pt}
    \item Benchmarking and comparing NLP attacks from previous works on standardized models \& datasets.
    \item Fast development of NLP attack methods by re-using abundant available modules.
    \item Performing ablation studies on individual components of proposed attacks and data augmentation methods.
    \item Training a model (CNN, LSTM, BERT, RoBERTa, etc.) on an augmented dataset.
    \item Adversarial training with attacks from the literature to improve a model's robustness.
\end{itemize}

\section{The \texttt{TextAttack} Framework}\label{sec:attack-components}

\texttt{TextAttack} aims to implement attacks which, given an NLP model, find a perturbation of an input sequence that satisfies the attack's goal and adheres to certain linguistic constraints. In this way, attacking an NLP model can be framed as a combinatorial search problem. The attacker must search within all potential transformations to find a sequence of transformations that generate a successful adversarial example. 

Each attack can be constructed from four components:
\begin{enumerate}[noitemsep,topsep=0pt]
\item A task-specific \textbf{goal function} that determines whether the attack is successful in terms of the model outputs. \\ Examples: untargeted classification, targeted classification, non-overlapping output, minimum BLEU score.
\item A set of \textbf{constraints} that determine if a perturbation is valid with respect to the original input. \\ Examples: maximum word embedding distance, part-of-speech consistency, grammar checker, minimum sentence encoding cosine similarity.
\item A \textbf{transformation} that, given an input, generates a set of potential perturbations. \\ Examples: word embedding word swap, thesaurus word swap, homoglyph character substitution.
\item A \textbf{search method} that successively queries the model and selects promising perturbations from a set of transformations. \\ Examples: greedy with word importance ranking, beam search, genetic algorithm.
\end{enumerate}

See \ref{appendix:four-components} for a full explanation of each goal function, constraint, transformation, and search method that's built-in to \texttt{TextAttack.}

\section{Developing NLP Attacks with \texttt{TextAttack}}
\label{sec:develop}
\vspace{-2mm}

\begin{figure}[t]
\centering
\includegraphics[width=.5\textwidth]{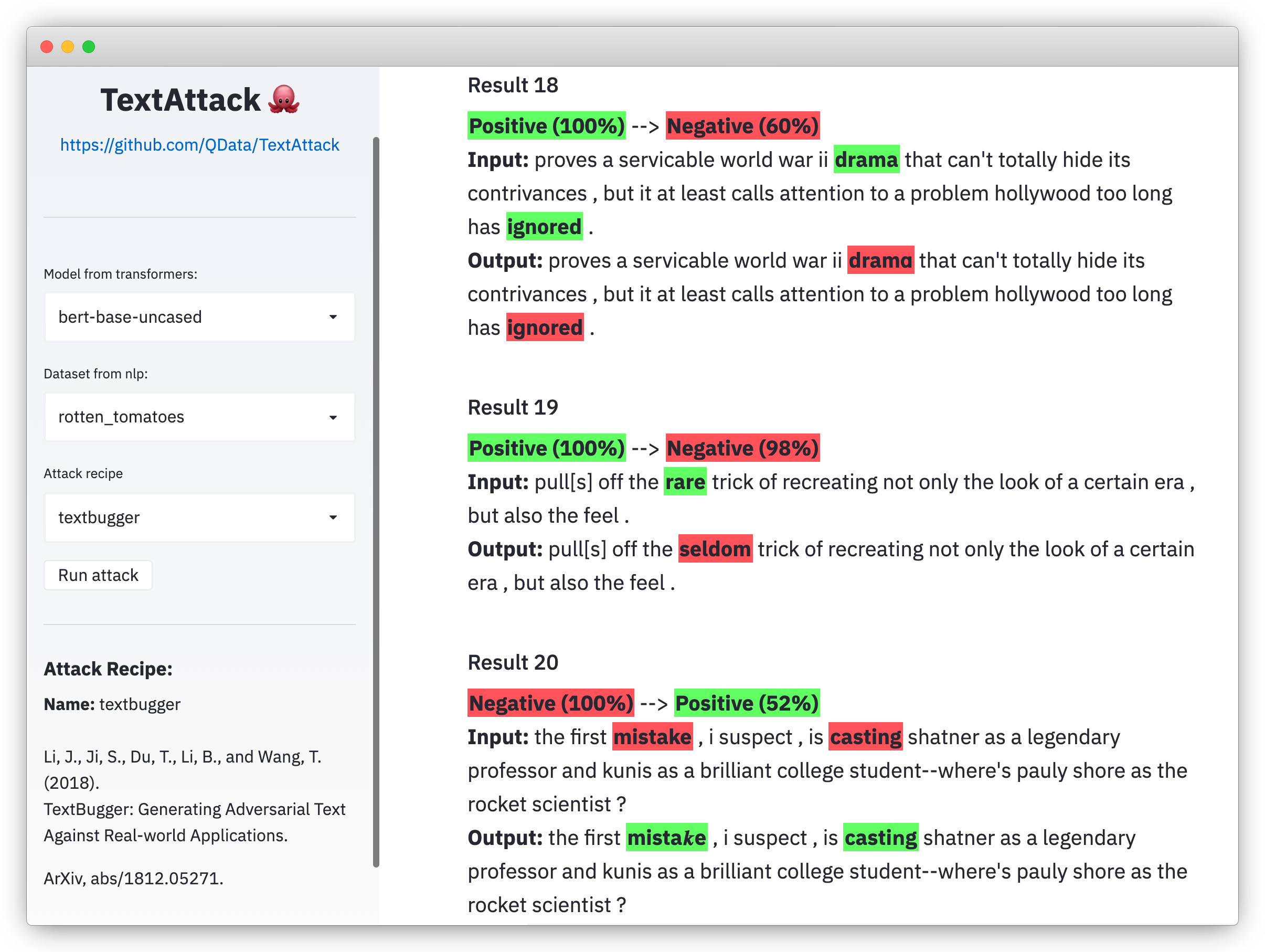}
\caption{Screenshot of \texttt{TextAttack}'s web interface running the TextBugger black-box attack \cite{Li2019TextBuggerGA}. \label{fig:web-demo}}
\end{figure}

\texttt{TextAttack} is available as a Python package installed from PyPI, or via direct download from GitHub. \texttt{TextAttack} is also available for use through our demo web app, displayed in Figure \ref{fig:web-demo}.

Python users can test attacks by creating and manipulating \texttt{Attack} objects. The command-line API offers \texttt{textattack attack}, which allows users to specify attacks from their four components or from a single attack recipe and test them on different models and datasets.

\texttt{TextAttack} supports several different output formats for attack results:
\begin{itemize}[topsep=0pt, partopsep=0pt]
  \setlength\itemsep{0em}
  \setlength{\parskip}{0pt}
    \item Printing results to stdout.
    \item Printing to a text file or CSV.
    \item Printing attack results to an HTML table.
    \item Writing a table of attack results to a visualization server, like Visdom or Weights \& Biases.
\end{itemize}

\subsection{Benchmarking Existing Attacks with Attack Recipes}
\texttt{TextAttack}'s modular design allows us to implement many different attacks from past work in a shared framework, often by adding only one or two new components. Table \ref{table:categorized-past-work} categorizes 16 attacks based on their goal functions, constraints, transformations and search methods.

All of these attacks are implemented as "attack recipes" in \texttt{TextAttack} and can be benchmarked with just a single command. See \ref{appendix:reproduction-results} for a comparison between papers' reported attack results and the results achieved by running \texttt{TextAttack}.

\subsection{Creating New Attacks by Combining Novel and Existing Components}

As is clear from Table \ref{table:categorized-past-work}, many components are shared between NLP attacks. New attacks often re-use components from past work, adding one or two novel pieces. \texttt{TextAttack} allows researchers to focus on the generation of new components rather than replicating past results. For example, \citet{Jin2019TextFooler} introduced \texttt{TextFooler} as a method for attacking classification and entailment models. If a researcher wished to experiment with applying \texttt{TextFooler}'s search method, transformations, and constraints to attack translation models, all they need is to implement a translation goal function in \texttt{TextAttack}. They would then be able to plug in this goal function to create a novel attack that could be used to analyze translation models.

\subsection{Evaluating Attacks on \texttt{TextAttack}'s Pre-Trained Models}

As of the date of this submission, \texttt{TextAttack} provides users with 82 pre-trained models, including word-level LSTM, word-level CNN, BERT, and other transformer based models pre-trained on various datasets provided by HuggingFace \href{https://github.com/huggingface/nlp/}{nlp}. Since \texttt{TextAttack} is integrated with the  \href{https://github.com/huggingface/nlp/}{nlp} library, it can automatically load the test or validation data set for the corresponding pre-trained model. While the literature has mainly focused on classification and entailment, \texttt{TextAttack}'s pretrained models enable research on the robustness of models across all GLUE tasks.
\begin{table*}
    \centering
    \scalebox{0.9}{
    \begin{tabular}{|>{\raggedright\arraybackslash}p{3.6cm}|>{\raggedright\arraybackslash}p{2.2cm}|>{\raggedright\arraybackslash}p{3.8cm}|>{\raggedright\arraybackslash}p{3.8cm}|>{\raggedright\arraybackslash}p{2.6cm}|}
     \hline
     \textbf{Attack Recipe} & \textbf{Goal Function} & \textbf{Constraints} & \textbf{Transformation} & \textbf{Search Method} \\
     \hline
     \texttt{bae} \newline \cite{garg2020bae} & Untargeted Classification & USE sentence encoding cosine similarity & BERT Masked Token Prediction & Greedy-WIR \\
     \hline
     \texttt{bert-attack} \newline \cite{li2020bertattack} & Untargeted Classification & USE sentence encoding cosine similarity, Maximum number of words perturbed & BERT Masked Token Prediction (with subword expansion) & Greedy-WIR \\
     \hline
     \texttt{deepwordbug} \newline \cite{Gao2018BlackBoxGO} & \{Untargeted, Targeted\} Classification & Levenshtein edit distance & \{Character Insertion, Character Deletion, Neighboring Character Swap, Character Substitution\}*  & Greedy-WIR \\
     \hline
     \texttt{alzantot,}\newline \texttt{fast-alzantot} \newline
     \cite{Alzantot2018GeneratingNL,Jia2019CertifiedRT} & Untargeted \{Classification, Entailment\}& Percentage of words perturbed, Language Model perplexity, Word embedding distance & Counter-fitted word embedding swap  & Genetic Algorithm  \\
     \hline
     \texttt{iga} \newline
     \cite{iga-wang2019natural} & Untargeted \{Classification, Entailment\} & Percentage of words perturbed, Word embedding distance & Counter-fitted word embedding swap  & Genetic Algorithm \\
     \hline
     \texttt{input-reduction} \newline \cite{feng2018pathologies} & Input Reduction & & Word deletion & Greedy-WIR \\
     \hline
     \texttt{kuleshov} \newline \cite{Kuleshov2018AdversarialEF} & Untargeted Classification & Thought vector encoding cosine similarity, Language model similarity probability & Counter-fitted word embedding swap & Greedy word swap \\
     \hline
     \texttt{hotflip} (word swap) \newline \cite{Ebrahimi2017HotFlipWA} & Untargeted Classification & Word Embedding Cosine Similarity, Part-of-speech match, Number of words perturbed & Gradient-Based Word Swap & Beam search \\
     \hline
     \texttt{morpheus} \newline \cite{morpheus-tan-etal-2020-morphin} & Minimum BLEU Score & & Inflection Word Swap & Greedy search \\
     \hline
     \texttt{pruthi} \newline \cite{pruthi2019combating} & Untargeted Classification & Minimum word length, Maximum number of words perturbed & \{Neighboring Character Swap, Character Deletion, Character Insertion, Keyboard-Based Character Swap\}* & Greedy search \\
     
     \hline
     \texttt{pso} \newline \cite{pso-zang-etal-2020-word} & Untargeted Classification & & HowNet Word Swap & Particle Swarm Optimization \\
     \hline
     \texttt{pwws} \newline \cite{pwws-ren-etal-2019-generating} & Untargeted Classification & & WordNet-based synonym swap & Greedy-WIR (saliency) \\
     \hline
     \texttt{seq2sick} (black-box) \newline \cite{cheng2018seq2sick} & Non-overlapping output & & Counter-fitted word embedding swap & Greedy-WIR \\
     \hline
     \texttt{textbugger} (black-box) \newline \cite{Li2019TextBuggerGA} & Untargeted Classification & USE sentence encoding cosine similarity & \{Character Insertion, Character Deletion, Neighboring Character Swap, Character Substitution\}* &  Greedy-WIR \\
     \hline
     \texttt{textfooler} \newline \cite{Jin2019TextFooler} & Untargeted \{Classification, Entailment\} & Word Embedding Distance, Part-of-speech match, USE sentence encoding cosine similarity & Counter-fitted word embedding swap & Greedy-WIR \\
    \hline
    \end{tabular}}
    \caption{\texttt{TextAttack} attack recipes categorized within our framework: search method, transformation, goal function, constraints. All attack recipes include an additional constraint which disallows the replacement of stopwords. Greedy search with Word Importance Ranking is abbreviated as Greedy-WIR. \newline \textit{* indicates a combination of multiple transformations}}
    \label{table:categorized-past-work}
\end{table*}
\FloatBarrier
\section{Utilizing \texttt{TextAttack} to Improve NLP Models}

\subsection{Evaluating Robustness of Custom Models}
\texttt{TextAttack} is model-agnostic - meaning it can run attacks on models implemented in any deep learning framework. Model objects must be able to take a string (or list of strings) and return an output that can be processed by the goal function. For example, machine translation models take a list of strings as input and produce a list of strings as output. Classification and entailment models return an array of scores. As long as the user's model meets this specification, the model is fit to use with \texttt{TextAttack}.

\subsection{Model Training}
\texttt{TextAttack} users can train standard LSTM, CNN, and transformer based models, or a user-customized model on any dataset from the \texttt{nlp} library using the \texttt{textattack train} command. 
Just like pre-trained models, user-trained models are compatible with commands like \texttt{textattack attack} and \texttt{textattack eval}.

\subsection{Data Augmentation}

While searching for adversarial examples, \texttt{TextAttack}'s transformations generate perturbations of the input text, and apply constraints to verify their validity. These tools can be reused to dramatically expand the training dataset by introducing perturbed versions of existing samples. The \texttt{textattack augment} command gives users access to a number of pre-packaged recipes for augmenting their dataset. This is a stand-alone feature that can be used with any model or training framework. When using \texttt{TextAttack}'s models and training pipeline, \texttt{textattack train --augment} automatically expands the dataset before training begins. Users can specify the fraction of each input that should be modified and how many additional versions of each example to create. This makes it easy to use existing augmentation recipes on different models and datasets, and is a great way to benchmark new techniques. 

\begin{figure}
\centering
    \includegraphics[width=.5\textwidth]{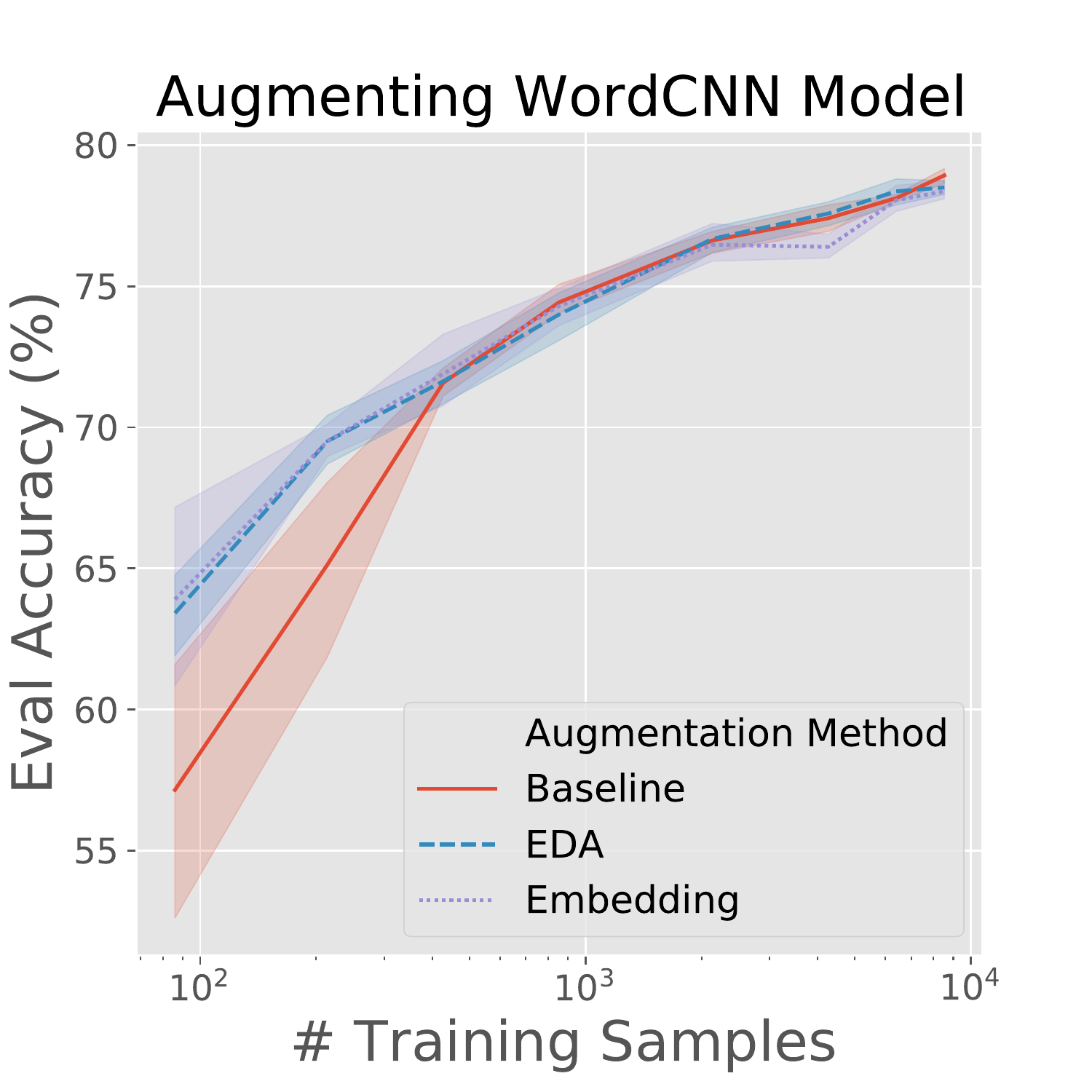}
    \caption{Performance of the built-in \texttt{WordCNN} model on the \texttt{rotten\_tomatoes} dataset with increasing training set size. Data augmentation recipes like \texttt{EasyDataAugmenter} (EDA, \cite{EDAWei2019}) and \texttt{Embedding} are most helpful when working with very few samples. Shaded regions represent $95\%$ confidence intervals over $N=5$ runs.}
    \label{fig:data_aug}
\end{figure}

Figure \ref{fig:data_aug} shows empirical results we obtained using \texttt{TextAttack}'s augmentation. Augmentation with \texttt{TextAttack} immediately improves the performance of a \texttt{WordCNN} model on small datasets.

\subsection{Adversarial Training}

\begin{table*}[]
\centering
 \scalebox{0.9}{\begin{tabular}{|l|l|c|c|c|c|c|}
 \cline{2-7}
 \multicolumn{1}{c|}{}& \multicolumn{6}{|c|}{\textbf{Attacked By}} \\ \hline
\textbf{Trained Against}  & \texttt{-}  & \texttt{deepwordbug} & \texttt{textfooler}  & \texttt{pruthi}  & \texttt{hotflip} & \texttt{bae}\\ \hline
baseline (early stopping)& \textbf{77.30\%} & 23.46\%& 2.23\% & 59.01\% & 64.57\% & 25.51\% \\
\texttt{deepwordbug} (20 epochs) &76.38\% & \textbf{35.07\%} & 4.78\% & 57.08\% & 65.06\% & 27.63\% \\
\texttt{deepwordbug} (75 epochs) &73.16\% & \textbf{44.74\%} & 13.42\%& 58.28\% & 66.87\% & 32.77\% \\
\texttt{textfooler} (20 epochs) & 61.85\% & 40.09\%& \textbf{29.63\%} & 52.60\% & 55.75\% & 39.36\% \\ \hline
\end{tabular}}
\caption{The default LSTM model trained on $3$k samples from the \texttt{sst2} dataset. The baseline uses early stopping on a clean training set. \texttt{deepwordbug} and \texttt{textfooler} attacks are used for adversarial training. `Accuracy Under Attack` on the eval set is reported for several different attack types. \label{table:adv_train}}
\end{table*}

With \texttt{textattack train --attack}, attack recipes can be used to create new training sets of adversarial examples. After training for a number of epochs on the clean training set, the attack generates an adversarial version of each input. This perturbed version of the dataset is substituted for the original, and is periodically regenerated according to the model's current weaknesses. The resulting model can be significantly more robust against the attack used during training. Table \ref{table:adv_train} shows the accuracy of a standard LSTM classifier with and without adversarial training against different attack recipes implemented in \texttt{TextAttack}.

\section{\texttt{TextAttack} Under the Hood}\label{sec:new}

\texttt{TextAttack} is optimized under-the-hood to make implementing and running adversarial attacks simple and fast.

\textbf{AttackedText.} A common problem with implementations of NLP attacks is that the original text is discarded after tokenization; thus, the transformation is performed on the tokenized version of the text. This causes issues with capitalization and word segmentation. Sometimes attacks swap a piece of a word for a complete word (for example, transforming \texttt{``aren't"} into \texttt{``aren'too"}). 

To solve this problem, \texttt{TextAttack} stores each input as a \texttt{AttackedText} object which contains the original text and helper methods for transforming the text while retaining tokenization. Instead of strings or tensors, classes in \texttt{TextAttack} operate primarily on \texttt{AttackedText} objects. When words are added, swapped, or deleted, an \texttt{AttackedText} can maintain proper punctuation and capitalization. The \texttt{AttackedText} also contains implementations for common linguistic functions like splitting text into words, splitting text into sentences, and part-of-speech tagging.

\textbf{Caching.} Search methods frequently encounter the same input at different points in the search. In these cases, it is wise to pre-store values to avoid unnecessary computation. For each input examined during the attack, \texttt{TextAttack} caches its model output, as well as the whether or not it passed all of the constraints. For some search methods, this memoization can save a significant amount of time.\footnote{Caching alone speeds up the genetic algorithm of \citet{Alzantot2018GeneratingNL} by a factor of 5.}

\section{Related Work}

We draw inspiration from the \texttt{Transformers} library \cite{Wolf2019TransformersSN} as an example of a well-designed Natural Language Processing library. Some of \texttt{TextAttack}'s models and tokenizers are implemented using \texttt{Transformers}.

\texttt{cleverhans} \cite{papernot2018cleverhans} is a library for constructing adversarial examples for computer vision models. Like \texttt{cleverhans}, we aim to provide methods that generate adversarial examples across a variety of models and datasets. In some sense, \texttt{TextAttack} strives to be a solution like \texttt{cleverhans} for the NLP community. Like \texttt{cleverhans}, attacks in \texttt{TextAttack} all implement a base \texttt{Attack} class. However, while \texttt{cleverhans} implements many disparate attacks in separate modules, \texttt{\texttt{TextAttack}} builds attacks from a library of shared components.

There are some existing open-source libraries related to adversarial examples in NLP. \texttt{Trickster}  proposes a method for attacking NLP models based on graph search, but lacks the ability to ensure that generated examples satisfy a given constraint \cite{TricksterKulynychHST18}. \texttt{TEAPOT} is a library for evaluating adversarial perturbations on text, but only supports the application of ngram-based comparisons for evaluating attacks on machine translation models \cite{TEAPOTMichel19Eval}. Most recently, \texttt{AllenNLP Interpret}  includes functionality for running adversarial attacks on NLP models, but is intended only for the purpose of interpretability, and only supports attacks via input-reduction or greedy gradient-based word swap \cite{Wallace2019AllenNLPIA}. \texttt{TextAttack} has a broader scope than any of these libraries: it is designed to be extendable to any NLP attack. 
\section{Conclusion}

We presented \texttt{TextAttack}, an open-source framework for testing the robustness of NLP models. \texttt{TextAttack} defines an attack in four modules: a goal function, a list of constraints, a transformation, and a search method. This allows us to compose attacks from previous work from these modules and compare them in a shared environment. These attacks can be reused for data augmentation and adversarial training. As new attacks are developed, we will add their components to \texttt{TextAttack}. We hope \texttt{TextAttack} helps lower the barrier to entry for research into robustness and data augmentation in NLP.

\section{Acknowledgements} The authors would like to thank everyone who has contributed to make \texttt{TextAttack} a reality: Hanyu Liu, Kevin Ivey, Bill Zhang, and Alan Zheng, to name a few. Thanks to the IGA creators \cite{iga-wang2019natural} for contributing an implementation of their algorithm to our framework. Thanks to the folks at HuggingFace for creating such easy-to-use software; without them, \texttt{TextAttack} would not be what it is today.

\bibliography{bib}
\bibliographystyle{acl_natbib}

\clearpage
\appendix
\section{Appendix}

\subsection{\texttt{TextAttack} in Five Lines or Less}
\label{appendix:five-lines-or-less}

 Table \ref{appendix:five-lines-or-less-table} provides some examples of tasks that can be accomplished in bash or Python with five lines of code or fewer. Note that every action has to be prefaced with a single line of code (\texttt{pip install textattack}).

\begin{table*}[bt!]
\scalebox{0.76}{
\begin{tabular}{p{3cm}|p{7cm}|p{9cm}}
\toprule
 & Task & Command \\ \midrule
\multirow{5}{*}{Run an attack} & TextFooler on an LSTM trained on the MR sentiment classification dataset & \texttt{textattack attack --recipe textfooler --model bert-base-uncased-mr --num-examples 100} \\ 
\cmidrule(l){2-3} 
 & TextFooler against BERT fine-tuned on SST-2 & \texttt{textattack attack --model bert-base-uncased-sst2 --recipe textfooler --num-examples 10} \\ 
\cmidrule(l){2-3} 
 & DeepWordBug on DistilBERT trained on the Quora Question Pairs paraphrase identification dataset: & \texttt{textattack attack --model distilbert-base-uncased-qqp --recipe deepwordbug --num-examples 100} 
 \\ \cmidrule(l){2-3} 
 & seq2sick (black-box) against T5 fine-tuned for English-German translation: & \texttt{textattack attack --model t5-en-de --recipe seq2sick --num-examples 100} 
 \\ \cmidrule(l){2-3} 
 & Beam search with beam width 4 and word embedding transformation and untargeted goal function on an LSTM: & \texttt{textattack attack --model lstm-mr --num-examples 20   --search-method beam-search:beam\_width=4 --transformation word-swap-embedding   --constraints repeat stopword max-words-perturbed:max\_num\_words=2 embedding:min\_cos\_sim=0.8 part-of-speech  --goal-function untargeted-classification} 
 
 \\ \midrule
\multirow{2}{*}{Data augmentation} & Augment dataset from 'examples.csv' using the EmbeddingAugmenter, swapping out 4\% of words, with 2 augmentations for example, withholding the original samples from the output CSV & \texttt{textattack augment --csv examples.csv --input-column text --recipe embedding --pct-words-to-swap 4 --transformations-per-example 2 --exclude-original} 
\\ \cmidrule(l){2-3}
 & \multirow{1}{*}{Augment a list of strings in Python} & \texttt{from textattack.augmentation import EmbeddingAugmenter} \\ 
 & & \texttt{augmenter = EmbeddingAugmenter() }\\ 
 & &\texttt{s = 'What I cannot create, I do not understand.'} \\
 & & \texttt{augmenter.augment(s)} \\  
 \midrule
\multirow{2}{*}{Train a model} & Train the default LSTM for 50 epochs on the Yelp Polarity dataset & \texttt{textattack train --model lstm --dataset yelp\_polarity --batch-size 64 --epochs 50 --learning-rate 1e-5}
\\ \cmidrule(l){2-3} 
 & Fine-tune bert-base on the CoLA dataset for 5 epochs & \texttt{textattack train --model bert-base-uncased --dataset glue:cola --batch-size 32 --epochs 5} 
 \\ \cmidrule(l){2-3} 
 & Fine-tune RoBERTa on the Rotten Tomatoes Movie Review dataset, first augmenting each example with 4 augmentations produced by the EasyDataAugmentation augmenter & \texttt{textattack train --model roberta-base --batch-size 64 --epochs 50 --learning-rate 1e-5 --dataset rotten\_tomatoes --augment eda --pct-words-to-swap .1 --transformations-per-example 4} 
 \\ \cmidrule(l){2-3} 
& Adversarially fine-tune DistilBERT on AG News using the HotFlip word-based attack, first training for 2 epochs on the original dataset &   \texttt{textattack train --model distilbert-base-cased --dataset ag\_news --attack hotflip --num-clean-epochs 2}
\\ \bottomrule
\end{tabular}%
}
\caption{With \texttt{TextAttack}, adversarial attacks, data augmentation, and adversarial training can be achieved in just a few lines of Bash or Python. \label{appendix:five-lines-or-less-table}}
\end{table*}

\subsection{Components of \texttt{TextAttack}}
\label{appendix:four-components}

\begin{figure*}[t]
\includegraphics[width=\linewidth]{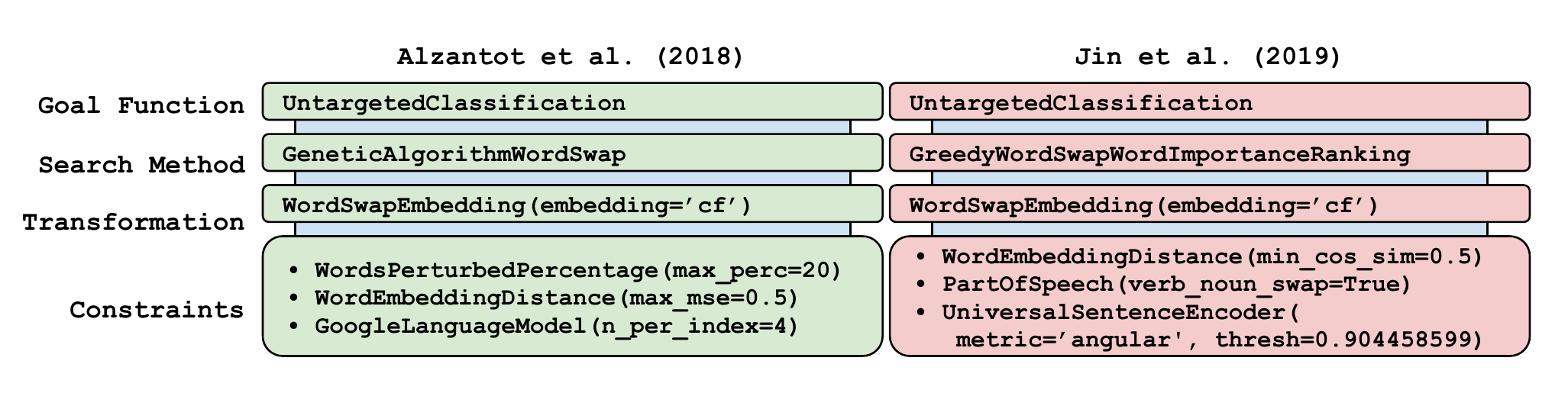}
\caption{\texttt{TextAttack} builds NLP attacks from a goal function, search method, transformation, and list of constraints. This shows attacks from \citet{Alzantot2018GeneratingNL} and \citet{Jin2019TextFooler} created using \texttt{TextAttack} modules.} 
\label{appendix:recipe-categorized-attacks}
\end{figure*}

This section explains each of the four components of the \texttt{TextAttack} framework and describes the components that are currently implemented. Figure \ref{appendix:recipe-categorized-attacks} shows the decomposition of two popular attacks \cite{Alzantot2018GeneratingNL,Jin2019TextFooler}.

\subsubsection{Goal Functions}

A goal function takes an input $x'$ and determines if it is satisfies the conditions for a successful attack in respect to the original input $x$.
Goal functions vary by task. For example, for a classification task, a successful adversarial attack could be changing the model's output to be a certain label. Goal functions also scores how "good" the given $x'$ is for achieving the desired goal, and this score can be used by the search method as a heuristic for finding the optimal solution.

\texttt{TextAttack} includes the following goal functions:
\begin{itemize}[topsep=0pt, partopsep=0pt]
  \setlength\itemsep{0em}
  \setlength{\parskip}{0pt}
  \item Untargeted Classification: Minimize the score of the correct classification label.
  \item Targeted Classification: Maximize the score of a chosen incorrect classification label.
  \item Input Reduction (Classification): Reduce the input text to as few wordas as possible while maintaining the same predicted label.
  \item Non-Overlapping Output (Text-to-Text): Change the output text such that no words in it overlap with the original output text.
  \item Minimzing BLEU Score (Text-to-Text): Change the output text such that the BLEU score between it and the original output text is minimized \cite{Papineni2001BleuAM}.
 \end{itemize}

\subsubsection{Constraints}

A perturbed text is only considered valid if it satisfies each of the attack's constraints. \texttt{TextAttack} contains four classes of constraints.

\paragraph{Pre-transformation Constraints}~
These constraints are used to preemptively limit how \texttt{x} can be perturbed and are applied before \texttt{x} is perturbed.
\begin{itemize}[topsep=0pt, partopsep=0pt]
  \setlength\itemsep{0em}
  \setlength{\parskip}{0pt}
  \item Stopword Modification: stopwords cannot be perturbed.
  \item Repeat Modification: words that have been already perturbed cannot be perturbed again.
  \item Minimum Word Length: words less than a certain length cannot be perturbed.
  \item Max Word Index Modification: words past a certain index cannot be perturbed.
  \item Input Column Modification: for tasks such as textual entailment where input might be composed of two parts (e.g. hypothesis and premise), we can limit which part we can transform (e.g. hypothesis). 
 \end{itemize}

\paragraph{Overlap}~
We measure the overlap between \texttt{x} and \texttt{x\_adv} using the following metrics on the character level and require it to be lower than a certain threshold as a constraint:

\begin{itemize}[topsep=0pt, partopsep=0pt]
  \setlength\itemsep{0em}
  \setlength{\parskip}{0pt}

\item Maximum BLEU score difference \cite{Papineni2001BleuAM}

\item Maximum chrF score difference \cite{Popovic2015chrFCN}

\item Maximum METEOR score difference \cite{Agarwal2008MeteorMA}

\item Maximum Levenshtein edit distance

\item Maximum percentage of words changed
\end{itemize}

\paragraph{Grammaticality}~
These constraints are typically intended to prevent the attack from creating perturbations which introduce grammatical errors. \texttt{TextAttack} currently supports the following constraints on grammaticality:

\begin{itemize}[topsep=0pt, partopsep=0pt]
  \setlength\itemsep{0em}
  \setlength{\parskip}{0pt}
\item Maximum number of grammatical errors induced, as measured by LanguageTool \cite{LanguageToolNaber2003rule}
\item Part-of-speech consistency: the replacement word should have the same part-of-speech as the original word. Supports taggers provided by flair, SpaCy, and NLTK.
\item Filtering out words that do not fit within the context based on the following language models:
    \begin{itemize}[topsep=0pt, partopsep=0pt]
      \setlength\itemsep{0em}
      \setlength{\parskip}{0pt}
        \item Google 1-billion words language model \cite{Jzefowicz2016ExploringTL}
        \item Learning To Write Language Model \cite{holtzman-etal-2018-learning-to-write} (as used by \cite{Jia2019CertifiedRT})
        \item GPT-2 language model \cite{radford2019GPT2}
   \end{itemize}
\end{itemize}

\paragraph{Semantics}~
Some constraints attempt to preserve semantics  between \texttt{x} and \texttt{x\_adv}. \texttt{TextAttack} currently provides the following built-in semantic constraints:

\begin{itemize}[topsep=0pt, partopsep=0pt]
  \setlength\itemsep{0em}
  \setlength{\parskip}{0pt}

\item Maximum swapped word embedding distance (or minimum cosine similarity)

\item Minimum cosine similarity score of sentence representations obtained by well-trained sentence encoders:
    \begin{itemize}[topsep=0pt, partopsep=0pt]
      \setlength\itemsep{0em}
      \setlength{\parskip}{0pt}
    \item Skip-Thought Vectors \cite{Kiros2015SkipThoughtV}
    \item Universal Sentence Encoder \cite{Cer2018UniversalSE}
    \item InferSent \cite{Conneau2017SupervisedLOInferSent}
    \item BERT trained for semantic similarity \cite{reimers-2019-sentence-bert}
    \end{itemize}

\item Minimum BERTScore \cite{bert-score2020}
\end{itemize}

\subsubsection{Transformations}
\label{sec:transformations}

A transformation takes an input and returns a set of potential perturbations. The transformation is agnostic of goal function and constraint(s): it returns all potential transformations.

We categorize transformations into two kinds: white-box and black-box.  \textit{White-box transformations} have access to the model and can query it or examine its parameters to help determine the transformation. For example, \citet{Ebrahimi2017HotFlipWA} determines potential replacement words based on the gradient of the one-hot input vector at the position of the swap. 
\textit{Black-box transformations} determine the potential perturbations without any knowledge of the model.

\texttt{TextAttack} currently supports the following transformations:
\begin{itemize}[topsep=0pt, partopsep=0pt]
  \setlength\itemsep{0em}
  \setlength{\parskip}{0pt}

\item Word swap with nearest neighbors in the counter-fitted embedding space \citep{mrkvsic2016counter}

\item WordNet word swap \cite{WordNetMiller1990IntroductionTW}

\item Word swap proposed by a masked language model \cite{garg2020bae,li2020bertattack}

\item Word swap gradient-based: swap word with another word in the vocabulary that maximize the model's loss \cite{Ebrahimi2017HotFlipWA} (white-box)

\item Word swap with characters transformed \cite{Gao2018BlackBoxGO}:
\begin{itemize}[topsep=0pt, partopsep=0pt]
  \setlength\itemsep{0em}
  \setlength{\parskip}{0pt}
        \item Character deleted
        \item Neighboring characters swapped
        \item Random character inserted
        \item Substituted with a random character
        \item Character substituted with a homoglyph
        \item Character substituted with a neighboring character from the keyboard \cite{pruthi2019combating}
    \end{itemize}

\item Word deletion

\item Word swap with another word in the vocabulary that has the same Part-of-Speech and sememe, where the sememe is obtained by HowNet~\citep{dong2006hownet}.

\item Composite transformation: returns the results of multiple transformations

\end{itemize}

\subsubsection{Search Methods}

The search method aims to find a perturbation that achieves the goal and satisfies all constraints. Many combinatorial search methods have been proposed for this process. \texttt{TextAttack} has implemented a selection of the most popular ones from the literature:

\begin{itemize}
\item \textbf{Greedy Search with Word Importance Ranking.} Rank all words according to some ranking function. Swap words one at a time in order of decreasing importance.

\item \textbf{Beam Search.} Initially score all possible transformations. Take the top $b$ transformations (where $b$ is a hyperparameter known as the "beam width") and iterate, looking at potential transformations for all sequences in the beam.

\item \textbf{Greedy Search.} Initially score transformations at all positions in the input. Swap words, taking the highest-scoring transformations first. (This can be seen as a case of beam search where $b=1$).

\item \textbf{Genetic Algorithm.} An implementation of the algorithm proposed by \citet{Alzantot2018GeneratingNL}. Iteratively alters the population through greedy perturbation of each population member and crossover between population numbers, with preference to the more successful members of the population. (We also support an alternate version, the "Improved Genetic Algorithm" proposed by \citet{iga-wang2019natural}).

\item \textbf{Particle Swarm Optimization.} A population-based evolutionary computation paradigms~\citep{kennedy1995particle} that exploits a population of interacting individuals to iteratively search for the optimal solution in the specific space~\citep{pso-zang-etal-2020-word}. The population is called a \textit{swarm} and individual agents are called \textit{particles}. Each particle has a position in the search space and moves with an adaptable \textit{velocity}.

\end{itemize}
\subsection{\texttt{TextAttack} Attack Reproduction Results}
\label{appendix:reproduction-results}

Table \ref{table:benchmark-comparison} displays a comparison of results achieved when running attacks in \texttt{TextAttack} alongside numbers reported in the original paper. All \texttt{TextAttack} benchmarks were run on pre-trained models provided by the library and can be reproduced in a single \texttt{textattack attack} command. There are a few important implementation differences:

\begin{itemize}
  \item The genetic algorithm benchmark comes from the faster genetic algorithm of \cite{jia-liang-2017-adversarial}. As opposed to the original algorithm of \cite{Alzantot2018GeneratingNL}, this implementation uses a fast language model, so it can query contexts of up to 5 words. Additionally, perplexity is compared to that of the original word, not the previous perturbation. Since these are more rigorous linguistic constraints, a lower attack success rate is expected.
  \item The LSTM models from BAE \cite{garg2020bae} were trained using counter-fitted GLoVe embeddings. The LSTM models from \texttt{TextAttack} were trained using normal GLoVe embeddings. Our models are consequently less robust to counter-fitted embedding synonym swaps, and a higher attack success rate is expected.
  \item The HowNet synonym set used in \texttt{TextAttack}'s PSO implementation is a concatenation of the three synonym sets used in the paper. This is necessary since \texttt{TextAttack} is dataset-agnostic and cannot expect to provide a set of synonyms for every possible dataset. Since the attack has more synonyms to choose from, \texttt{TextAttack}'s PSO implementation is slightly more successful. 
\end{itemize}

\begin{table*}[]
\resizebox{\textwidth}{!}{
\begin{tabular}{llcccc|ccccc}
\toprule
\multicolumn{1}{c}{}    && \multicolumn{4}{c|}{\textbf{LSTM}}     & \multicolumn{5}{c}{\textbf{BERT-Base}} \\
\multicolumn{1}{c}{}    && \textbf{MR}    & \textbf{SST-2} & \textbf{IMDB} & \textbf{AG}    & \textbf{MR}    & \textbf{SST-2} & \textbf{IMDB}& \textbf{SNLI}  & \textbf{AG}    \\ \midrule
& Reported   & -     & -    & 97.0 / 14.7   & -     & -     & -    & -    & -     & -     \\
\multirow{-2}{*}{\texttt{alzantot} \cite{Alzantot2018GeneratingNL}}     & \texttt{TextAttack} & 64.6 / 17.8 & 70.8 / 18.3& 73.0 / 4.0  & 27.7 / 11.6    & 40.7 / 19.1 & 46.5 / 20.7& 46.7 / 7.3 & 74.9 / 12.3 & 18.1 / 12.6 \\
& Reported   & 70.2 / -    & -    & 73.2 / -    & -     & 48.3 / -    & -    & 45.9 / -   & -     & -     \\
\multirow{-2}{*}{\texttt{bae} \cite{garg2020bae}}   & \texttt{TextAttack} & 74.4 / 12.3 & 72.7 / 13.5& 88.8 / 2.6  & 21.4 / 6.3  & 61.5 / 15.2 & 66.6 / 14.5& 55.6 / 3.2 & 78.4 / 7.1  & 16.9 / 7.4  \\
& Reported   & -     & -    & -     & 72.5 / -    & -     & -    & -    & -     & -     \\
\multirow{-2}{*}{\texttt{deepwordbug} \cite{Gao2018BlackBoxGO}} & \texttt{TextAttack} & 86.3 / 16.8 & 82.6 / 17.1& 97.6 / 5.2  & 83.4 / 19.4 & 78.2 / 21.2 & 81.3 / 18.9& 80.9 / 5.3 & 99.0 / 9.8  & 60.7 / 25.1 \\
& Reported   & -     & 93.8 / 9.1 & 100.0 / 3.7 & -     & -     & 91.2 / 8.2 & 98.7 / 3.7 & 78.9 / 11.7 & -     \\
\multirow{-2}{*}{\texttt{pso} \cite{pso-zang-etal-2020-word}}   & \texttt{TextAttack} & 94.9 / 10.7 & 96.5 / 11.5& 100.0 / 1.3 & 83.7 / 12.7 & 92.7 / 11.9 & 91.3 / 12.9& 100.0 / 1.2& 91.8 / 6.2  & 79.4 / 16.7 \\
& Reported   & 96.2 / 14.9 & -    & 99.7 / 5.1     & 95.8 / 18.6 & 86.7 / 16.7 & -    & 85.0 / 6.1 & 95.5 / 18.5 & 86.7 / 22.0 \\
\multirow{-2}{*}{\texttt{textfooler} \cite{Jin2019TextFooler}}  & \texttt{TextAttack} & 97.4 / 13.6    & 98.8 / 14.2& 100.0 / 2.4 & 95.3 / 17.2 & 88.7 / 18.7    & 94.8 / 16.9& 100.0 / 7.2& 96.3 / 7.2  & 79.5 / 23.5   \\ \bottomrule
\end{tabular}}
\caption{Comparison between our re-implemented attacks and the original source code in terms of success rate (left number) and percentage of perturbed words (right number). Numbers that are not found in the literature are marked as ``-''. 1000 samples are randomly selected for evaluation from all these datasets except IMDB (100 samples are used for IMDB since some attack methods like Genetic and PSO take over 4 days to finish 1000 samples).}
\label{table:benchmark-comparison}
\end{table*}

\subsection{\texttt{TextAttack} Attack Prototypes}
\label{appendix:attack-prototypes}

This section displays ``attack prototypes'' for each attack recipe implemented in \texttt{TextAttack}. This is a concise way to print out the components of a given attack along with its parameters. These are directly copied from the output of running \texttt{TextAttack}.

\textbf{Alzantot Genetic Algorithm \cite{Alzantot2018GeneratingNL}}
\begin{lstlisting}[caption={},label={list:usage},captionpos=b, basicstyle=\scriptsize, frame=lrtb]
Attack(
  (search_method): GeneticAlgorithm(
    (pop_size):  60
    (max_iters):  20
    (temp):  0.3
    (give_up_if_no_improvement):  False
  )
  (goal_function):  UntargetedClassification
  (transformation):  WordSwapEmbedding(
    (max_candidates):  8
    (embedding_type):  paragramcf
  )
  (constraints):
    (0): MaxWordsPerturbed(
        (max_percent):  0.2
        (compare_against_original):  True
      )
    (1): WordEmbeddingDistance(
        (embedding_type):  paragramcf
        (max_mse_dist):  0.5
        (cased):  False
        (include_unknown_words):  True
        (compare_against_original):  False
      )
    (2): GoogleLanguageModel(
        (top_n):  None
        (top_n_per_index):  4
        (compare_against_original):  False
      )
    (3): RepeatModification
    (4): StopwordModification
    (5): InputColumnModification(
        (matching_column_labels):  ['premise', 'hypothesis']
        (columns_to_ignore):  {'premise'}
      )
  (is_black_box):  True
)
\end{lstlisting}

\textbf{Alzantot Genetic Algorithm (faster) \cite{Jia2019CertifiedRT}}
\begin{lstlisting}[caption={},label={list:usage},captionpos=b, basicstyle=\scriptsize, frame=lrtb]

Attack(
  (search_method): GeneticAlgorithm(
    (pop_size):  60
    (max_iters):  20
    (temp):  0.3
    (give_up_if_no_improvement):  False
  )
  (goal_function):  UntargetedClassification
  (transformation):  WordSwapEmbedding(
    (max_candidates):  8
    (embedding_type):  paragramcf
  )
  (constraints): 
    (0): MaxWordsPerturbed(
        (max_percent):  0.2
      )
    (1): WordEmbeddingDistance(
        (embedding_type):  paragramcf
        (max_mse_dist):  0.5
        (cased):  False
        (include_unknown_words):  True
      )
    (2): LearningToWriteLanguageModel(
        (max_log_prob_diff):  5.0
      )
    (3): RepeatModification
    (4): StopwordModification
  (is_black_box):  True
)
\end{lstlisting}

\textbf{BAE} \cite{garg2020bae}
\begin{lstlisting}[caption={},label={list:usage},captionpos=b, basicstyle=\scriptsize, frame=lrtb]
Attack(
  (search_method): GreedyWordSwapWIR(
    (wir_method):  delete
  )
  (goal_function):  UntargetedClassification
  (transformation):  WordSwapMaskedLM(
    (method):  bae
    (masked_lm_name):  bert-base-uncased
    (max_length):  256
    (max_candidates):  50
  )
  (constraints):
    (0): PartOfSpeech(
        (tagger_type):  nltk
        (tagset):  universal
        (allow_verb_noun_swap):  True
        (compare_against_original):  True
      )
    (1): UniversalSentenceEncoder(
        (metric):  cosine
        (threshold):  0.936338023
        (window_size):  15
        (skip_text_shorter_than_window):  True
        (compare_against_original):  True
      )
    (2): RepeatModification
    (3): StopwordModification
  (is_black_box):  True
)
\end{lstlisting}

\textbf{BERT-Attack \cite{li2020bertattack}}
\begin{lstlisting}[caption={},label={list:usage},captionpos=b, basicstyle=\scriptsize, frame=lrtb]
Attack(
  (search_method): GreedyWordSwapWIR(
    (wir_method):  unk
  )
  (goal_function):  UntargetedClassification
  (transformation):  WordSwapMaskedLM(
    (method):  bert-attack
    (masked_lm_name):  bert-base-uncased
    (max_length):  256
    (max_candidates):  48
  )
  (constraints):
    (0): MaxWordsPerturbed(
        (max_percent):  0.4
        (compare_against_original):  True
      )
    (1): UniversalSentenceEncoder(
        (metric):  cosine
        (threshold):  0.2
        (window_size):  inf
        (skip_text_shorter_than_window):  False
        (compare_against_original):  True
      )
    (2): RepeatModification
    (3): StopwordModification
  (is_black_box):  True
)
\end{lstlisting}

\textbf{DeepWordBug \cite{Gao2018BlackBoxGO}}
\begin{lstlisting}[caption={},label={list:usage},captionpos=b, basicstyle=\scriptsize, frame=lrtb]
Attack(
  (search_method): GreedyWordSwapWIR(
    (wir_method):  unk
  )
  (goal_function):  UntargetedClassification
  (transformation):  CompositeTransformation(
    (0): WordSwapNeighboringCharacterSwap(
        (random_one):  True
      )
    (1): WordSwapRandomCharacterSubstitution(
        (random_one):  True
      )
    (2): WordSwapRandomCharacterDeletion(
        (random_one):  True
      )
    (3): WordSwapRandomCharacterInsertion(
        (random_one):  True
      )
    )
  (constraints):
    (0): LevenshteinEditDistance(
        (max_edit_distance):  30
        (compare_against_original):  True
      )
    (1): RepeatModification
    (2): StopwordModification
  (is_black_box):  True
)
\end{lstlisting}

\textbf{HotFlip \cite{Ebrahimi2017HotFlipWA}}
\begin{lstlisting}[caption={},label={list:usage},captionpos=b, basicstyle=\scriptsize, frame=lrtb]
Attack(
  (search_method): BeamSearch(
    (beam_width):  10
  )
  (goal_function):  UntargetedClassification
  (transformation):  WordSwapGradientBased(
    (top_n):  1
  )
  (constraints):
    (0): MaxWordsPerturbed(
        (max_num_words):  2
        (compare_against_original):  True
      )
    (1): WordEmbeddingDistance(
        (embedding_type):  paragramcf
        (min_cos_sim):  0.8
        (cased):  False
        (include_unknown_words):  True
        (compare_against_original):  True
      )
    (2): PartOfSpeech(
        (tagger_type):  nltk
        (tagset):  universal
        (allow_verb_noun_swap):  True
        (compare_against_original):  True
      )
    (3): RepeatModification
    (4): StopwordModification
  (is_black_box):  False
)
\end{lstlisting}

\textbf{Input Reduction \cite{feng2018pathologies}}
\begin{lstlisting}[caption={},label={list:usage},captionpos=b, basicstyle=\scriptsize, frame=lrtb]
Attack(
  (search_method): GreedyWordSwapWIR(
    (wir_method):  delete
  )
  (goal_function):  InputReduction(
    (maximizable):  True
  )
  (transformation):  WordDeletion
  (constraints):
    (0): RepeatModification
    (1): StopwordModification
  (is_black_box):  True
)
\end{lstlisting}

\textbf{Kuleshov \cite{Kuleshov2018AdversarialEF}}
\begin{lstlisting}[caption={},label={list:usage},captionpos=b, basicstyle=\scriptsize, frame=lrtb]
Attack(
  (search_method): GreedySearch
  (goal_function):  UntargetedClassification
  (transformation):  WordSwapEmbedding(
    (max_candidates):  15
    (embedding_type):  paragramcf
  )
  (constraints):
    (0): MaxWordsPerturbed(
        (max_percent):  0.5
        (compare_against_original):  True
      )
    (1): ThoughtVector(
        (embedding_type):  paragramcf
        (metric):  max_euclidean
        (threshold):  -0.2
        (window_size):  inf
        (skip_text_shorter_than_window):  False
        (compare_against_original):  True
      )
    (2): GPT2(
        (max_log_prob_diff):  2.0
        (compare_against_original):  True
      )
    (3): RepeatModification
    (4): StopwordModification
  (is_black_box):  True
)
\end{lstlisting}

\textbf{MORPHEUS \cite{morpheus-tan-etal-2020-morphin}}
\begin{lstlisting}[caption={},label={list:usage},captionpos=b, basicstyle=\scriptsize, frame=lrtb]
Attack(
  (search_method): GreedySearch
  (goal_function):  MinimizeBleu(
    (maximizable):  False
    (target_bleu):  0.0
  )
  (transformation):  WordSwapInflections
  (constraints):
    (0): RepeatModification
    (1): StopwordModification
  (is_black_box):  True
)
\end{lstlisting}

\textbf{Particle Swarm Optimization \cite{pso-zang-etal-2020-word}}
\begin{lstlisting}[caption={},label={list:usage},captionpos=b, basicstyle=\scriptsize, frame=lrtb]
Attack(
  (search_method): ParticleSwarmOptimization
  (goal_function):  UntargetedClassification
  (transformation):  WordSwapHowNet(
    (max_candidates):  -1
  )
  (constraints):
    (0): RepeatModification
    (1): StopwordModification
    (2): InputColumnModification(
        (matching_column_labels):  ['premise', 'hypothesis']
        (columns_to_ignore):  {'premise'}
      )
  (is_black_box):  True
)
\end{lstlisting}

\textbf{Pruthi Keyboard Char-Swap Attack \cite{pruthi2019combating}}
\begin{lstlisting}[caption={},label={list:usage},captionpos=b, basicstyle=\scriptsize, frame=lrtb]
Attack(
  (search_method): GreedySearch
  (goal_function):  UntargetedClassification
  (transformation):  CompositeTransformation(
    (0): WordSwapNeighboringCharacterSwap(
        (random_one):  False
      )
    (1): WordSwapRandomCharacterDeletion(
        (random_one):  False
      )
    (2): WordSwapRandomCharacterInsertion(
        (random_one):  False
      )
    (3): WordSwapQWERTY
    )
  (constraints):
    (0): MaxWordsPerturbed(
        (max_num_words):  1
        (compare_against_original):  True
      )
    (1): MinWordLength
    (2): StopwordModification
    (3): RepeatModification
  (is_black_box):  True
)
\end{lstlisting}

\textbf{PWWS \cite{pwws-ren-etal-2019-generating}}
\begin{lstlisting}[caption={},label={list:usage},captionpos=b, basicstyle=\scriptsize, frame=lrtb]
Attack(
  (search_method): GreedyWordSwapWIR(
    (wir_method):  pwws
  )
  (goal_function):  UntargetedClassification
  (transformation):  WordSwapWordNet
  (constraints): 
    (0): RepeatModification
    (1): StopwordModification
  (is_black_box):  True
)
\end{lstlisting}

\textbf{seq2sick \cite{cheng2018seq2sick}}
\begin{lstlisting}[caption={},label={list:usage},captionpos=b, basicstyle=\scriptsize, frame=lrtb]
Attack(
  (search_method): GreedyWordSwapWIR(
    (wir_method):  unk
  )
  (goal_function):  NonOverlappingOutput
  (transformation):  WordSwapEmbedding(
    (max_candidates):  50
    (embedding_type):  paragramcf
  )
  (constraints):
    (0): LevenshteinEditDistance(
        (max_edit_distance):  30
        (compare_against_original):  True
      )
    (1): RepeatModification
    (2): StopwordModification
  (is_black_box):  True
)
\end{lstlisting}

\textbf{TextBugger \cite{Li2019TextBuggerGA}}
\begin{lstlisting}[caption={},label={list:usage},captionpos=b, basicstyle=\scriptsize, frame=lrtb]
Attack(
  (search_method): GreedyWordSwapWIR(
    (wir_method):  unk
  )
  (goal_function):  UntargetedClassification
  (transformation):  CompositeTransformation(
    (0): WordSwapRandomCharacterInsertion(
        (random_one):  True
      )
    (1): WordSwapRandomCharacterDeletion(
        (random_one):  True
      )
    (2): WordSwapNeighboringCharacterSwap(
        (random_one):  True
      )
    (3): WordSwapHomoglyphSwap
    (4): WordSwapEmbedding(
        (max_candidates):  5
        (embedding_type):  paragramcf
      )
    )
  (constraints):
    (0): UniversalSentenceEncoder(
        (metric):  angular
        (threshold):  0.8
        (window_size):  inf
        (skip_text_shorter_than_window):  False
        (compare_against_original):  True
      )
    (1): RepeatModification
    (2): StopwordModification
  (is_black_box):  True
)
\end{lstlisting}

\textbf{TextFooler} \cite{Jin2019TextFooler}
\begin{lstlisting}[caption={},label={list:usage},captionpos=b, basicstyle=\scriptsize, frame=lrtb]
Attack(
  (search_method): GreedyWordSwapWIR(
    (wir_method):  del
  )
  (goal_function):  UntargetedClassification
  (transformation):  WordSwapEmbedding(
    (max_candidates):  50
    (embedding_type):  paragramcf
  )
  (constraints):
    (0): WordEmbeddingDistance(
        (embedding_type):  paragramcf
        (min_cos_sim):  0.5
        (cased):  False
        (include_unknown_words):  True
        (compare_against_original):  True
      )
    (1): PartOfSpeech(
        (tagger_type):  nltk
        (tagset):  universal
        (allow_verb_noun_swap):  True
        (compare_against_original):  True
      )
    (2): UniversalSentenceEncoder(
        (metric):  angular
        (threshold):  0.840845057
        (window_size):  15
        (skip_text_shorter_than_window):  True
        (compare_against_original):  False
      )
    (3): RepeatModification
    (4): StopwordModification
    (5): InputColumnModification(
        (matching_column_labels):  ['premise', 'hypothesis']
        (columns_to_ignore):  {'premise'}
      )
  (is_black_box):  True
)
\end{lstlisting}

\clearpage
\appendix

\end{document}